\documentclass[12pt,A4]{article}
\usepackage[margin=1in,footskip=0.25in]{geometry}

\usepackage{graphicx, color}  
\usepackage{bm}
\usepackage{amsmath, amsfonts}

\usepackage{natbib}
\usepackage[noend]{algpseudocode}
\usepackage{algorithmicx,algorithm}
\usepackage{multirow}
\usepackage{subfigure} 
\usepackage{booktabs}
\usepackage{threeparttable}
\usepackage{hyperref}
\usepackage{enumerate}
\usepackage{ntheorem}

\hypersetup{
	colorlinks=true,
	linkcolor=black,
	filecolor=black,      
	urlcolor=black,
	citecolor=black,
}

\def\x{\bm{x}}   \def\z{\bm{z}}
\def\u{\bm{u}}   \def\v{\bm{v}} 
 \def\I{\bm{I}}
  
\def\U{\bm{U}} \def\V{\bm{V}}

  \def\M{\bm{M}}

\def \RRR {\mathbb{R}} \def \EEE {\mathbb{E}}

\begin{document}
	
	\title{Explainable Recommendation Systems by Generalized Additive Models with Manifest and Latent Interactions}
\author{Yifeng Guo$^1$, Yu Su$^2$, Zebin Yang$^1$, and Aijun Zhang$^{1,}$\thanks{Corresponding author. Email: ajzhang@umich.edu}\\
	{\normalsize  $^1$Department of Statistics and Actuarial Science, The University of Hong Kong,}\\
	{\normalsize Pokfulam Road, Hong Kong SAR, China}\\
	{\normalsize $^2$AI Lab, Suoxinda Holdings Limited, Shenzhen, Guangdong, China}}
\date{} 

\maketitle

\begin{abstract}
	In recent years, the field of recommendation systems has attracted increasing attention to developing predictive models that provide explanations of why an item is recommended to a user. The explanations can be either obtained by post-hoc diagnostics after fitting a relatively complex model or embedded into an intrinsically interpretable model. In this paper, we propose the explainable recommendation systems based on a generalized additive model with manifest and latent interactions (GAMMLI). This model architecture is intrinsically interpretable, as it additively consists of the user and item main effects, the manifest user-item interactions based on observed features, and the latent interaction effects from residuals. Unlike conventional collaborative filtering methods, the group effect of users and items are considered in GAMMLI. It is beneficial for enhancing the model interpretability, and can also facilitate the cold-start recommendation problem. A new Python package GAMMLI is developed for efficient model training and visualized interpretation of the results. By numerical experiments based on simulation data and real-world cases, the proposed method is shown to have advantages in both predictive performance and explainable recommendation. 
	
	\vskip 6.5pt \noindent {\bf Keywords}: 
	Generalized additive model; Interaction effects; Matrix factorization; Model interpretability; Personalized recommendation; Cold start problem. 
\end{abstract}

	\section{Introduction}
	\label{sec1}
	In the last decades, the recommendation system (RecSys) that predicts user preferences on multiple items has been widely adopted in many fields, including banking, the Internet, E-commerce, and social media. Traditional recommendation systems mainly fall into two categories: content-based and collaborative filtering (CF)-based recommendation. The former originates from information retrieval and information filtering~\cite{adomavicius2005toward}, and it models user and item profiles with assorted available content information. Two examples of content-based recommendation are the term frequency/inverse document frequency (TF-IDF) that specifies keyword weights~\citep{salton1989automatic} and the Fab system that recommends Web pages to users~\citep{balabanovic1997fab}. The CF-based recommendation systems are developed for matching users with similar tastes and preferences. User-based CF for the GroupLens news recommendation system is one of the earliest CF approaches by \cite{resnick1994grouplens} and later the item-based CF method is discussed by \cite{sarwar2001item}. Popularized by Netflix competition, the latent factor model based on matrix factorization (MF; \citealp{koren2009matrix}) became a widely known CF method with relatively high prediction performance. The RecSys research community has extended MF techniques in various aspects, including combining multiple matrices with network relationships among the variables \citep{zhang2011novel}, integrating MF with topic models \citep{wang2015collaborative} and integrating side information using matrix co-factorization technique \citep{pujahari2020pair}. Both content-based and CF-based methods have pros and cons. In particular, the CF techniques do not model the user or item features as carefully as the content-based methods. Combining these two types of methods leads to the hybrid recommendation system. Among others, the most successful hybrid is the factorization machine (FM; \citealp{rendle2010factorization,rendle2012factorization}) that extends matrix factorization for generic modeling of features. In recent years, the RecSys technique has been enhanced through deep learning in various ways, including wide and deep network \citep{cheng2016wide},  DeepFM \citep{guo2017deepfm}, neural collaborative filtering \citep{he2017neural} and multi-criteria collaborative filtering \citep{nassar2020novel} which all utilize the complex neural networks to richly express the user and item features and their dummy IDs.
	
	The state of the art recommendation systems includes MF, FM, and DeepFM reviewed above, which have superb predictive performance. However, they merely make recommendations, but can hardly tell why an item is recommended to a user in terms of user or item features.  In this sense, the MF, FM, and DeepFM methods are usually referred to as black-box models, which are known to be lacking transparency, persuasiveness, effectiveness, trustworthiness, and user satisfaction.
	There exist two categories of explainable recommendations in the literature. One is the post-hoc diagnostics that generates explanations after the recommendation has been made; see e.g., the explanation mining by \cite{peake2018explanation}. The other category is intrinsic embedding that tries to embed explicit interpretability into the recommendation models. Examples include the explicit factor model based on phrase-level sentiment analysis \citep{zhang2014explicit} and the neural attentional regression model with review-level explanations by \cite{chen2018neural}. Both examples are supplemented by user reviews on each item through text analysis. Such user-item response data together with textual review collection can be also modeled by the tensor analysis framework; see e.g., \cite{wang2018explainable}. A more comprehensive review of such sort of explainable recommendations can be referred to \cite{zhang2020explainable}.
	
	In this paper, we develop a new kind of intrinsically interpretable RecSys from the statistical modeling point of view, in particular, based on the idea of the generalized additive model (GAM; \citealp{hastie1990generalized}). 
	Consider a special case of RecSys problem with multiple users but only one item (either label or rating), then it corresponds to the standard machine learning problem. For data collected with user features $\x\in\RRR^p$ and a single item response $y\in \RRR$,  the GAM takes the following form
	\begin{equation}\label{GAM}
	g(\EEE[y]) = \mu + h_1(x_1) + \cdots + h_p(x_p),
	\end{equation}
	where $\mu$ is the intercept, $h(\cdot)$ is a smooth nonparametric function, and $g$ is a pre-specified link function. 
	In the fast-emerging area of interpretable machine learning,  GAM is known to process a good balance between prediction performance and inherent interpretability. Such a statistically interpretable model is a special case of additive index model (AIM) that has been successfully re-deployed into the explainable neural network (xNN) architectures \citep{vaughan2018explainable, yang2020enhancing} through univariate-input subnetwork parametrization of $h(\cdot)$ in (\ref{GAM}). It has been further extended to the GAMI-Net recently proposed by \cite{yang2020gami} to include structured pairwise interaction effects through bivariate-input subnetworks. Meanwhile, it draws our attention that GAM plays an essential role in the recent work of \cite{agarwal2020neural} under the rephrased name of neural additive models, which corresponds to a special case of xNN and GAMI-Net.
	
	The RecSys problem has multiple item responses per user, and the data collection includes user features $\x\in\RRR^p$, item features $\z\in\RRR^q$, and usually incomplete (user, item)-responses. It is our goal to find an appropriate way of applying GAM in (\ref{GAM}) for modeling such RecSys data of user-item features and responses. It is straightforward to model the main effects based on either user or item features. For the RecSys problem, the user-item cross interactions play critical roles in predicting the (user, item)-responses. We propose a two-way approach to model such user-item interactions, including an explicit way of modeling the manifest interactions induced from the observed user and item features, and an implicit way of modeling the latent interactions induced from other unobserved features. This leads to generalized additive models with manifest and latent interactions (GAMMLI) for explainable recommendation systems. Moreover, for pursuing enhanced model interpretability,  the latent interactions are modeled with user and item group structures. The details about the GAMMLI methodology and how to interpret GAMMLI results are presented in Section~\ref{gammli_methodology}. 
	
	The computational methods for GAMMLI estimation are presented in Section~\ref{computation_aspect}, where we discuss the stage-wise network training algorithm, latent interaction estimation, and hyperparameter tuning. In Section~\ref{experiments}, we conduct numerical experiments based on both synthetic data and three real-world cases.
	It is shown that the proposed method has not only high prediction accuracy but also superior model interpretability. Finally, Section~\ref{conclusion} concludes the paper with directions for future work.
	
	\section{GAMMLI Methodology} \label{gammli_methodology}
	Assume a RecSys data consists of $m$ users (indexed by $i=1,\ldots,m$) and $n$ items (indexed by $j=1,\ldots,n$). The observed data includes user features $\x_i\in\RRR^p$, item features $\z_j\in\RRR^q$, and (user, item)-responses by $y_{ij}$. Based on (\ref{GAM}), we propose a novel explainable RecSys framework which provides not only the intuitive interpretation of observed features but also the heuristic interpretation of unobserved features. GAMMLI consists of three components, i.e., main effects, manifest interactions, and latent interactions.	
	\begin{itemize}
		\item {\bf Main user and item effects:} the main effects can be modeled by (\ref{GAM}) based on user features $\x$ and item features $\z$.
		\item {\bf Manifest user-item interactions:} the manifest interaction effects $I^{(m)}_{ij}$ are induced by observed user features $\x$ and item features $\z$. They can be modeled explicitly by $I^{(m)}_{ij} = \sum_{a, b} h(x_{ia}, z_{jb})$ for selective $a\in \{1,\ldots, p\}$ and $b\in \{1, \ldots, q\}$.
		\item {\bf Latent user-item interactions:} the latent interaction effects $I^{(l)}_{ij}$ are induced by other unobserved user or item features. They can be modeled implicitly through low-rank approximation
		$I^{(l)}_{ij} \approx \sum_{c=1}^r u_{ic} v_{jc}$ for a relatively small integer $r$.
	\end{itemize}
	Putting these effects together, we have the GAMMLI model of the form
	\begin{equation}\label{GAMMLI}
	g(\EEE[y_{ij}])=\mu + \sum_{a=1}^p h_{a}^{(u)}(x_{i a})+\sum_{b= 1}^q h_{b}^{(v)}(z_{j b}) + \sum_{a, b}h^{(u,v)}_{a, b}(x_{i a},z_{j b})+\sum_{c=1}^r u_{ic} v_{jc},
	\end{equation}
	where the link function $g$ is specified according to different types of (user, item)-responses. 
	
	\begin{figure}[htb!]
		\centering
		\includegraphics[width=1.0\textwidth]{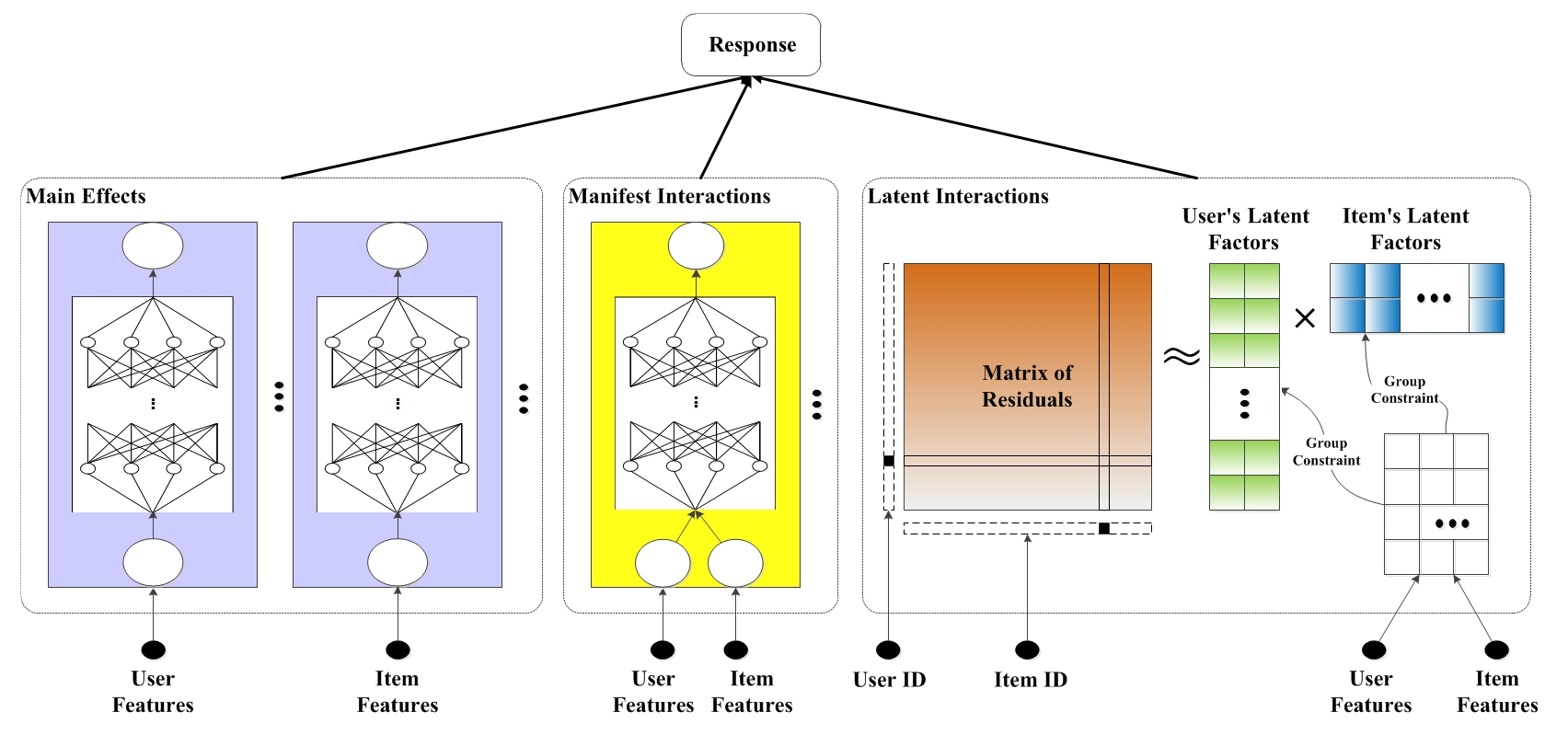}
		\caption{The network architecture of GAMMLI. In the left part, the network fits $p$ user main effects and $q$ item main effects based on explicitly observed features. In the middle part, the user-item manifest interactions are modeled by bivariate network inputs. The right part presents the group-constrained low-rank approximation for latent user-item interactions. These three modules are additively combined to form the final response.} \label{framework}
	\end{figure}
	
	The model architecture of GAMMLI is shown in Fig.~\ref{framework}. In particular, GAMMLI is fitted in a sequential way with three stages. In the first and second stages, the explicitly observed user and item features are used to model main effects and manifest user-item interactions.  In the third stage, the residuals from the first two stages are used to fit the latent user-item interactions. More details about such a sequential strategy are discussed in Section~3.
	
	\subsection{Main Effects and Manifest Interactions}
	Consider the three $h(\cdot)$ terms in (\ref{GAMMLI}), which are usually fitted by nonparametric methods like splines or trees. Recently, \citep{yang2020gami} proposed a GAMI-Net based on  additive subnetworks with better accuracy and interpretability than multiple existing machine learning methods. Such neural network parametrization is incorporated in this paper for modeling smooth $h(\cdot)$ functions in (\ref{GAMMLI}). Each main effect function, either $h_{a}^{(u)}(x_{a})$ of a user feature, or $h_{b}^{(v)}(z_{b})$ of an item feature, can be represented by a univariate-input subnetwork. Each user-item manifest interaction $h^{(u,v)}_{a, b}(x_{a},z_{b})$ can be represented by a bivariate-input subnetwork. For the sake of identifiability, each main effect or manifest interaction is assumed to have mean zero. 
	The categorical features can be modeled by bias nodes, where each bias node captures the intercept effect of a corresponding dummy variable.  More details about the subnetwork configuration can be referred to \cite{yang2020gami}. 
	
	Manifest user-item interactions are fitted to the residuals from the main effect prediction. 
	Each manifest interaction subnetwork consists of two input nodes, one for a user feature and the other for an item feature. 
	Note that if we directly use GAMI-Net \citep{yang2020gami}, all the user-user, user-item, and item-item interactions would be taken into account. However, in this paper, we restrict to the user-item cross manifest interactions only, since they are more interesting to model the interactions among user-item responses. On the other hand, we consider all two-way user-item manifest interactions without imposing weak heredity constraint or interaction filtering.

	
	\subsection{Latent Interaction Effects}
	Latent user-item interactions $I^{(l)}_{ij}$ are used to model the unobserved features after fitting the main effect and manifest interactions. Denote $\M_{m \times n}$ as the residual matrix with missing values. The observed entries are indexed by $\Omega = \{(i,j): \M_{ij} \mbox{ is not missing} \}$. It can be  approximated by low-rank matrix factorization as $I^{(l)}_{ij} \approx \sum_{c=1}^r u_{ic} v_{jc}$ for each user $i$  and each item $j$. Here, the column vectors $\{\u_c, \v_c\}$ for $c=1,\ldots,r$ are called user latent factors and item latent factors, respectively.  Using matrix notations, we can write them as $\U =[\u_1, \ldots, \u_r]$ of size $m\times r$ and $\V = [\v_1, \ldots, \v_r]$ of size $n\times r$.

	Unlike the existing matrix factorization methods, we take the group effect into consideration. The users and items are simultaneously clustered based on explicitly observed features with a biclustering technique.  The group constraint can increase the quality of recommendation in heterogeneous groups.  
	Denote by $S_k^{(u)}$ for $k=1,\ldots,K$ the $K$ disjoint clusters for the users, and by $S_l^{(v)}$ for $l=1,\ldots,L$ the $L$ disjoint clusters of the items. For each user-item pair $(i, j)$, let $k(i)$ be the corresponding cluster index for user $i$ and let $l(j)$ be the corresponding cluster index for item $j$. Then, we can find the cluster centroids  
	\begin{equation}\label{ClusterCentroid}
	\widetilde{\U}_{i\ast} =  \frac{1}{\big|S_{k(i)}^{(u)}\big|}\sum_{i'\in S_{k(i)}^{(u)}} \U_{i'\ast}  \ \mbox{ and } \ 
	\widetilde{\V}_{j\ast} =  \frac{1}{\big|S_{l(j)}^{(v)}\big|}\sum_{j'\in S_{l(j)}^{(v)}} \V_{j'\ast}, \quad\forall (i,j),
	\end{equation}
	so as to form the cluster centroid matrices $\widetilde{\U}$ and $\widetilde{\V}$. Then, we suggest to estimate the latent factors by the following regularized least squares, 
	\begin{equation}\label{latent_matrix}
	\min_{\U,\V} \|P_{\Omega}(\M- \U\V^T)\|_F^2+\lambda \left(\|\U-\widetilde{\U}\|_{F}^2+\|\V-\widetilde{\V}\|_{F}^2\right),
	\end{equation}
	where $P_{\Omega}$ is the projection of a matrix that preserves the entries indexed by $\Omega$, the subscript $F$ denotes the Frobenius norm of a matrix, and $\lambda\geq 0$ is the regularization strength.
	
	The problem (\ref{latent_matrix}) can be solved by alternating least squares similar to the SoftImpute algorithm \citep{mazumder2010spectral, hastie2015matrix}.  Given the pilot estimates for $\hat\U, \hat\V$, and pre-specified clusters, we may obtain the cluster centroid matrices by (\ref{ClusterCentroid}), then alternatively estimate $\V$ by 
	$$
	\min_{\V} \| P_{\Omega}(\M- \hat\U\V^T)\|_F^2 +\lambda \|\V-\widetilde{\V}\|_F^2,
	$$
	and estimate $\U$ with updated $\hat\V$ by 
	$$
	\min_{\U} \| P_{\Omega}(\M- \U\hat\V^T)\|_F^2 +\lambda \|\U-\widetilde{\U}\|_F^2,
	$$
	each corresponding to a multi-response ridge regression with closed-form solutions. A detailed procedure based on soft-imputed  singular value decomposition and alternative least squares is presented in Algorithm~\ref{alg1}. 	
	
	\begin{algorithm}[htb!]
		\caption{Latent Interactions-ALS} 
		\label{alg1}
		\begin{algorithmic}[1]
			\State Initialize residual matrix $\M$, $\hat\U=\U^{\ast}\bm{\Sigma}, \hat\V=\V^{\ast}\bm{\Sigma}$ where $\U^{\ast}$ is orthonormal matrix, $\bm{\Sigma}=\I_r$ is an identity matrix and $\V^{\ast}=0$. Based on the pre-specified clusters, calculate cluster centroid matrices $\widetilde{\U}, \widetilde{\V}$.
			\State  Given $\hat\U=\U^{\ast}\bm{\Sigma}, \hat\V=\V^{\ast}\bm{\Sigma}, \widetilde{\U}, \widetilde{\V}$, solve:
			$$
			\min_{\V} \| P_{\Omega}(\M- \hat\U\V^T)\|_F^2 +\lambda \|\V-\widetilde{\V}\|_F^2,
			$$
			\State Let $\ddot{\V}=\V-\widetilde{\V}$ and $\M^{\ast} = P_{\Omega}(\M-\hat\U \hat{\V}^T) + \hat\U(\hat\V-\widetilde{\V})^T$. We update $\hat\V$ by solving:
			$$
			\min_{\ddot{\V}} \|\M^{\ast} - \hat\U \ddot{\V}^T\|_F^2+\lambda \|\ddot{\V}\|_F^2,
			$$
			with solution
			$$
			\ddot{\V}^T=(\bm{\Sigma}^2 + \lambda \I)^{-1} \bm{\Sigma} {\U^{\ast}}^T \M^{\ast}.
			$$
			\State $\hat\V=\ddot{\V}+\widetilde{\V}$. Obtain SVD result for $\hat\V \bm{\Sigma}$ and update by the following steps:
			\begin{enumerate}[i]
				\item update $\V^{\ast} \gets$ left-singular vectors;
				\item update $\bm{\Sigma} \gets$ singular values;
				\item update $\hat\V \gets \V^{\ast}\bm{\Sigma}$ and cluster centroid matrix $\widetilde{\V}$.
			\end{enumerate}
			
			\State By symmetry, solve for $\U$ with solution
			$$
			\ddot\U=\M^{\ast} \V^{\ast} \bm{\Sigma} (\bm{\Sigma}^2 + \lambda \I)^{-1}.
			$$
			\State $\hat\U=\ddot{\U}+\widetilde{\U}$. Obtain SVD result for $\hat\U \bm{\Sigma}$ and update by the following steps:
			\begin{enumerate}[i]
				\item update $\U^{\ast} \gets$ left-singular vectors;
				\item update $\bm{\Sigma} \gets$ singular values;
				\item update $\hat\U \gets \U^{\ast}\bm{\Sigma}$ and cluster centroid matrix $\widetilde{\U}$.
			\end{enumerate}
			\State Repeat 2-6 until convergence.
			\State \Return reconstruction matrix $\hat\M = \hat\U \hat\V^T$.
		\end{algorithmic}
	\end{algorithm}
	
	The group constraints are incorporated into (\ref{latent_matrix}) to enhance the interpretability of the proposed GAMMLI method for explainable RecSys, with particular benefits in solving cold-start problems. The cold-start problem targets brand new users or items without existing information about user-item responses, and they widely exist in the real RecSys applications. Cold-start problems are hard to be dealt with by collaborative filtering methods since there are no records to find similar patterns with other users or items. By the proposed GAMMLI framework, we can take advantage of the auxiliary features associated with new users or items. Such features can be used not only for computing main effects and manifest interactions but also for obtaining the specific cluster assignments. The cluster centroids can be used as the latent interactions for the cold-start cases.  Therefore in GAMMLI, the clustering procedure bridges the observed features with latent factors and provides a heuristic interpretation of latent interactions.
	
	\subsection{GAMMLI Results Interpretation}
	It is straightforward to interpret the GAMMLI results by additive effect decomposition (\ref{GAMMLI}). We may characterize the contribution of each main effect and manifest interaction, as well as the contribution by the latent interactions. Besides, we can visualize the results.
	
	The importance ratio (IR) is a way to quantify the importance by the percentage of variation explained. For the modeled main effects and manifest interactions, their variations can be empirically evaluated by the sum of squared marginal predictions. For the modeled low-rank latent interactions, we can simply evaluate their sum of squares, too. Converting them into the percentage scale, we define the IR of each modeled effect as follows,
	\begin{equation}
	\begin{aligned}
	{\rm IR}(\hat{h}_{a}^{(u)}) &= \frac{1}{|\Omega|-1} \sum_{(i,j) \in \Omega} \hat{h}_a^{(u)}(x_{ia})^2/T\equiv  D(\hat{h}_{a}^{(u)})/T,\\
	{\rm IR}(\hat{h}_{b}^{(v)}) &= \frac{1}{|\Omega|-1} \sum_{(i,j) \in \Omega} \hat{h}_{b}^{(v)}(z_{jb})^2/T \equiv D(\hat{h}_b^{(v)})/T,\\
	{\rm IR}(\hat{h}^{(u,v)}_{a,b}) &= \frac{1}{|\Omega|-1} \sum_{(i,j) \in \Omega} \hat{h}^{(u,v)}_{a,b}(x_{ia},z_{jb})^2/ T \equiv D(\hat{h}^{(u,v)}_{a,b})/T,\\
	{\rm IR}(\{\hat\U, \hat\V\})&= \frac{1}{|\Omega|-1}\sum_{(i,j) \in \Omega} \Big(\hat\U_{i\ast}\hat\V_{j\ast}^T - \bar{M}\Big)^2/T  \equiv D(\{\hat\U, \hat\V\})/T,
	\end{aligned}\label{IR}
	\end{equation}
	where $|\Omega|$ denotes the total number of observed entries in the training data, $\bar{M}$ denotes the mean of estimated latent interactions and $T = \sum_{a=1}^p D(\hat{h}_a^{(u)}) + \sum_{b= 1}^q  D(\hat{h}_{b}^{(v)})+ \sum_{a,b} D(\hat{h}^{(u,v)}_{a,b}) + D(\{\hat\U, \hat\V\})$ represents the total variation.
	
	The partial relationship of the main effects and manifest interactions can be visualized by univariate line plots and bivariate heatmaps, similar to \cite{yang2020gami}. For the latent interactions subject to group structures, we can also use the heatmap to visualize the user-item cross effects. Examples of these plots are provided in Section~4. 
	
	\section{Computational Methods} \label{computation_aspect}
	The training of GAMMLI is conducted sequentially in three stages, including the estimations of main effects, manifest interactions, and latent interactions.
	
	\bigskip
	\textbf{Step 1:} In the main effects training, all main effect subnetworks are estimated simultaneously, and the user-item	interaction effects are fixed to zero. The trainable parameters in the network are updated by mini-batch gradient descent with adaptive learning rates determined by the Adam optimizer.  The main effects training step will stop when the maximum training epochs are reached or the given early stopping criterion is met. Each main effect is then centered to have a mean zero. The contribution of each main effect can be quantified by their variations as shown in (\ref{IR}). By selecting top-$s_1$ main effect subnetworks, nonsignificant main effects are accordingly removed. Note the optimal value of $s_{1}$ is automatically selected in the modeling process subject to the validation loss.
	
	\bigskip
	\textbf{Step 2:} After the main effects training finished, the subnetworks of pairwise user-item interactions are trained using the prediction residuals of the first stage. All user-item manifest interactions are taken into account due to the hierarchical principle.	Each estimated manifest interaction is centered to have mean zero, for which the offset is added to the bias node in the output layer. The manifest interactions training will stop similarly as in the first stage. The manifest interaction effects are also pruned by selecting the top-$s_2$ interaction effects according to $D(\hat{h}^{(u,v)}_{a,b})$ in (\ref{IR}) for selective $a\in \{1,\ldots, p\}$ and $b \in \{1, \ldots, q\}$. Also note there is no need to manually specify $s_{2}$, as it is automatically selected using the validation loss. Then, the main effects and manifest interactions will be jointly fine-tuned to improve performance. More details can be referred to~\cite{yang2020gami}.
	
	\bigskip
	\textbf{Step 3:} Before the latent interactions are estimated, all users and items will be assigned into $K$ and $L$ clusters, respectively. Various clustering algorithms have been proposed in the machine learning literature. And in fact, any clustering algorithm can be employed here. For simplicity, we just incorporate the popularly used K-means clustering. The clustering results, i.e., $S_k^{(u)}$ ($k=1,\ldots,K$) for users and $S_l^{(v)}$  ($l=1,\ldots,L$) for items will be used for the next step modeling.
	
	\bigskip
	\textbf{Step 4:} The prediction residuals of the first/second stage are used to fit latent interaction effects as shown in Algorithm \ref{alg1}, which  alternatively solves $\hat\U$ and $\hat\V$ by solving a multi-response ridge regression. For regression tasks, residuals are calculated by $y - \hat{y_{1}} - \hat{y_{2}}$. In classification scenarios, residuals are calculated as is in gradient boosting machine (GBM;\cite{friedman2001greedy}). That is, we first transform the binary response from $\{0,1\}$ to $\{-1,1\}$. The residual is then computed by  $\tilde{y}= 2y/(1+\exp(2y_i(\hat{y_1}+\hat{y_2})))$, where $y$ is true response, and $\hat{y_{1}}, \hat{y_{2}}$ are estimated responses in the first and second stages, respectively. Finally, The training of GAMMLI is summarized in Algorithm \ref{alg2}.
	
	\begin{algorithm}[htb!]
		\caption{GAMMLI Training} \label{alg2}
		\begin{algorithmic}[1]
			\State Train and select the top-$s_1$ main effect subnetworks.
			\State Train and select the top-$s_2$ manifest interaction subnetworks.
			\State Cluster (K-means) user and item features into $K$ and $L$ clusters, respectively.
			\State Train latent interaction effects via Algorithm \ref{alg1}.
		\end{algorithmic}
	\end{algorithm}
	
	\bigskip
	\textbf{Hyperparameter Tuning}. There exist several hyperparameters in GAMMLI, which are critical for the final prediction performance. The GAMMLI is configured with the following empirical settings. The subnetwork structure is configured to [20, 10] with hyperbolic tangent nodes. For the neural network training, the initial learning rate is set to 0.001, and the number of tuning epochs is fixed to 200. The epochs of training main effects, manifest interactions, and latent interactions are all set to 1000. A 20\% validation set is split for early stopping in each stage, and the early stopping threshold is set to 100 epochs. For interaction training, the rank of latent interactions $r$ is set to 3.
	
	Three important hyperparameters are identified in the latent interaction estimation stage, including users group number $K$, items group number $L$, and regularization strength $\lambda$. To pursue better performance, a coarse-to-fine grid-search method is employed to optimize $(K, L, \lambda)$. Specifically, the initial search space is set as $K$ (2 to 30), $L$ (2 to 30), and $\lambda$ (0 to 50). The search space is evenly partitioned such that each hyperparameter has 5 grid points. We enumerate all the possible 125 hyperparameter combinations and select the best performing one. Then, the second-iteration search is conducted sequentially by repeatedly centering on the best performing grid point, reduce the search space, and scatter high-granularity grid points. Such a procedure will be repeated and will be stopped after the fifth iteration. After hyperparameter optimization, a suitable combination of $(K, L, \lambda)$ can be obtained.

	\section{Numerical Experiments} \label{experiments}
	
	In this section, we evaluate our model via two simulation scenarios and three real datasets in offline settings. We begin by introducing the experimental setup and then report and analyze the experimental results.
	
	Four competitive benchmark models are introduced for comparison, including eXtreme Gradient Boosting (XGBoost), singular value decomposition (SVD), factorization machine (FM), and DeepFM. Four compared models cover different types of recommendation systems, i.e., machine learning model (XGBoost), collaborative filtering model (SVD), and complicated monolithic method (FM \& DeepFM).
	
	Each dataset is split into training, validation, and test sets. In SVD, we set the rank to 5. For both FM and DeepFM, we set different loss types for different tasks, i.e., mean squared error (MSE) for regression tasks and Log Loss for classification tasks. The number of training epochs is set differently for different tasks. In XGBoost, the maximum tree depth is selected from \{3,4,5,6,7,8\}. Other hyperparameters of benchmark models are set to their default values. All experiments are repeated 10 times, where we use different random train test split for modeling fitting and evaluation. For each dataset, we evaluate the test set root mean squares error (RMSE) and mean absolute error (MAE) for regression tasks and the area under the ROC curve (AUC) and Log Loss for binary classification tasks.
	
	All the experiments are implemented using the Python environment. The proposed GAMMLI is implemented using the \textit{TensorFlow 2.0} platform, with the open-sourced package uploaded to the Github\footnote{\href{https://github.com/SelfExplainML/GAMMLI}{https://github.com/SelfExplainML/GAMMLI}}. The SVD is implemented using {\tt surprise} package\footnote{\href{http://surpriselib.com/}{http://surpriselib.com/}}. FM and DeepFM are implemented by the TensorFlow package\footnote{\href{https://github.com/ChenglongChen/tensorflow-DeepFM}{https://github.com/ChenglongChen/tensorflow-DeepFM}} and XGBoost  is implemented by the popular {\tt xgboost} package\footnote{\href{https://github.com/dmlc/xgboost}{https://github.com/dmlc/xgboost}}.
	
	\subsection{Simulation Study}
	A synthetic data is first generated, where the numbers of users and items are set to $m=1000$ and $n=1000$, respectively. Each user or item has 5 numerical features and a latent vector with rank 3. The user/item features and latent vectors are generated from isotropic Gaussian blobs. The features are scaled within $[0,1]$ and latent vectors are scaled within $[-1,1]$. The actual numbers of user and item groups are both set to 10. To mimic the real data case where only a small percentage of responses is collected, the response matrix is generated as a sum of feature effect and latent interactions with size $m \times n$ and a missing rate $\omega=0.9$. A noise term generated from the standard normal distribution is also included when generating the response matrix. Both regression and classification scenarios will be included in the simulation study. 
	
	In the regression scenario, the feature effects of user and item are generated as follows:
	\begin{equation}
	\begin{aligned}
	y = 5x_1 + 5z_1^2 + 0.5e^{-4(z_2+x_3)+4} + 5sin(2\pi x_2 z_3) + 3 \sum_{k=1}^3 u_{k} v_{k}^T +\epsilon,
	\end{aligned}
	\label{eq7}
	\end{equation} 
	where $x,z$ represent user and item features and $\sum_{k=1}^3 u_{k} v_{k}^T$ denotes the sum of latent interactions. This case exactly follows the model formulation of GAMMLI. It can be observed that each user or item has two noisy features. As for the classification scenario, the formulation follows the regression setting but the response is binarized as $y = \mathbb{I}_{\{y > 0.5\}}$, where $\mathbb{I}$ is an indicator function. 
	
\begin{figure}[ht!]%
		\centering
		\subfigure[Regression]{\includegraphics[width=0.8\textwidth]{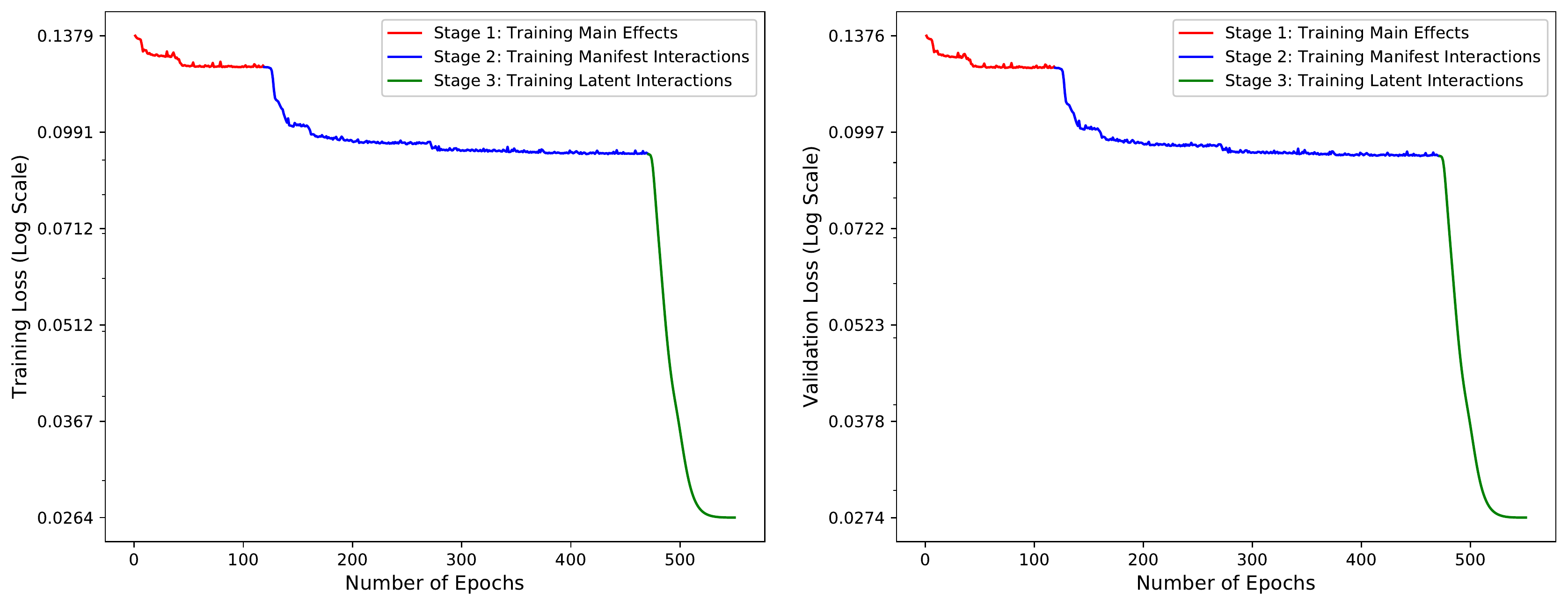}}
		\quad
		\subfigure[Classification]{\includegraphics[width=0.8\textwidth]{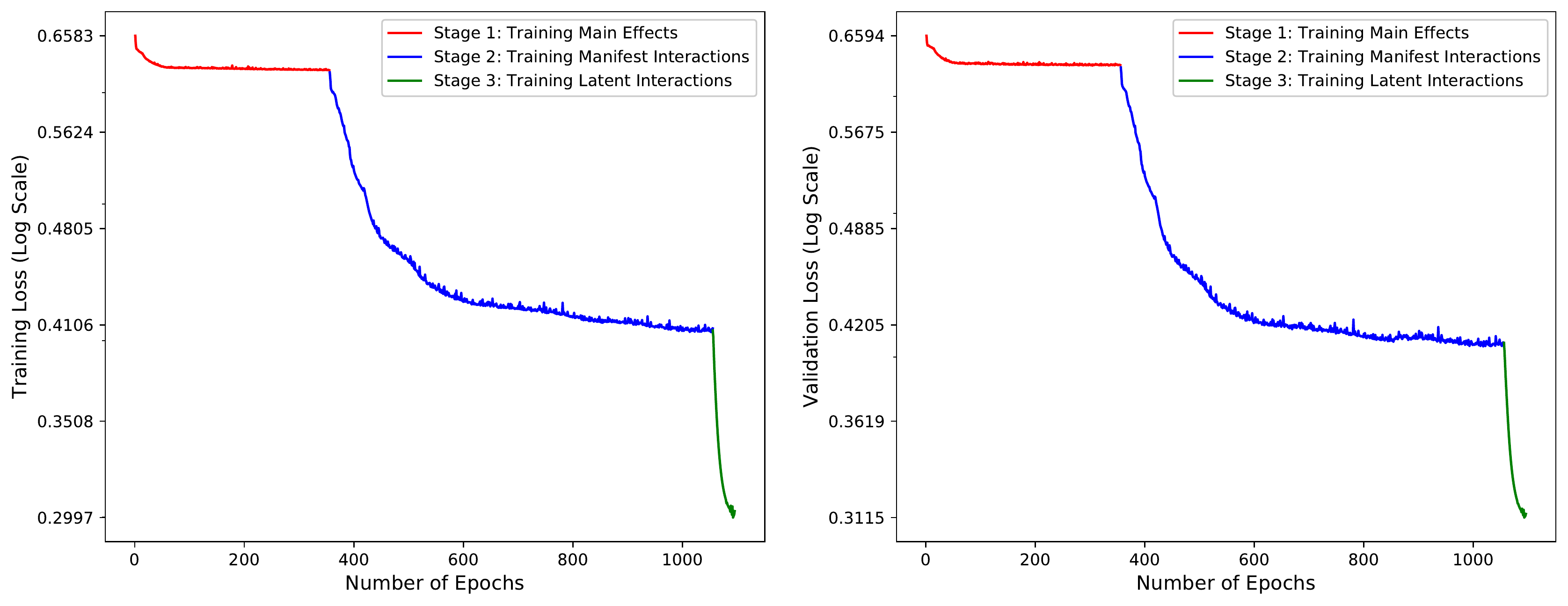}}
		\caption{Training and validation losses of the simulation study.}
		\label{simulation_trajectories}
	\end{figure}

	Fig.~\ref{simulation_trajectories} shows the training and validation losses of the simulation study. It can be observed that the losses drop significantly as manifest interactions and latent interactions are added to the model. Each component of GAMMLI contributes greatly to model accuracy.
	
\begin{figure}[!t]%
		\centering
		\subfigure[Ground truth]{\includegraphics[width=1.0\textwidth]{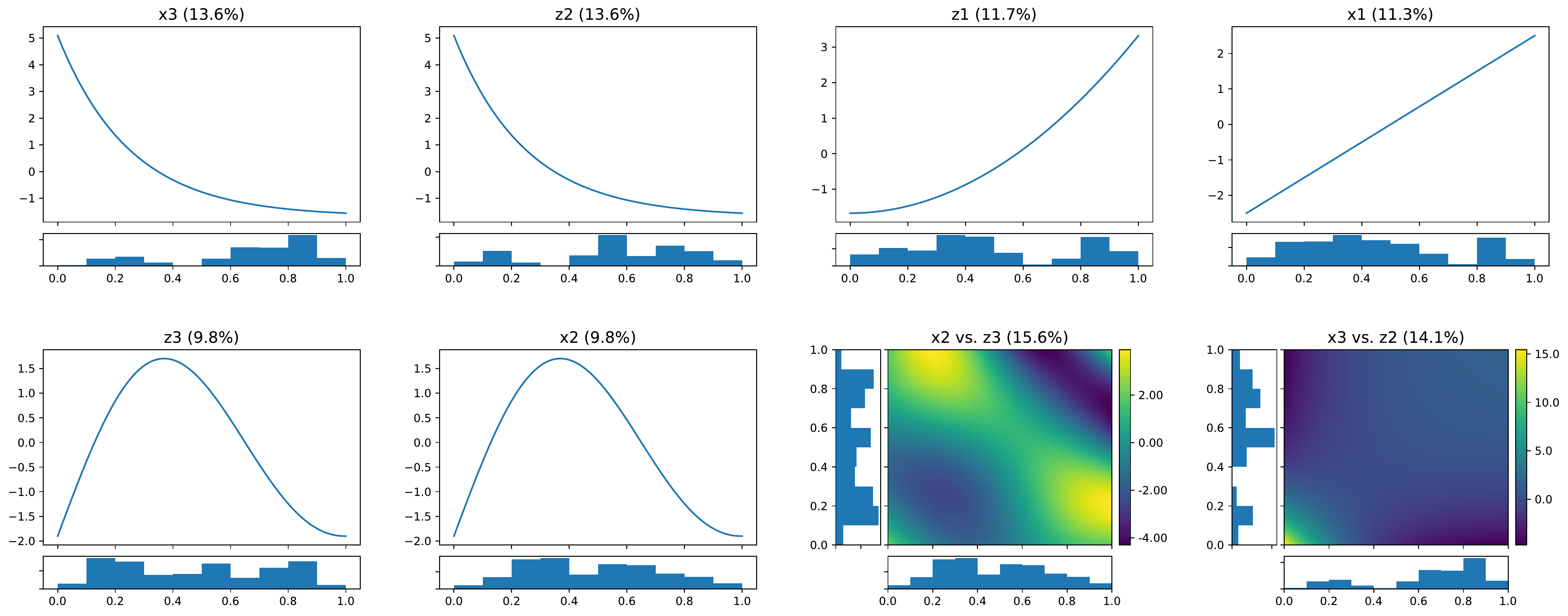}}
		\subfigure[GAMMLI]{\includegraphics[width=1.0\textwidth]{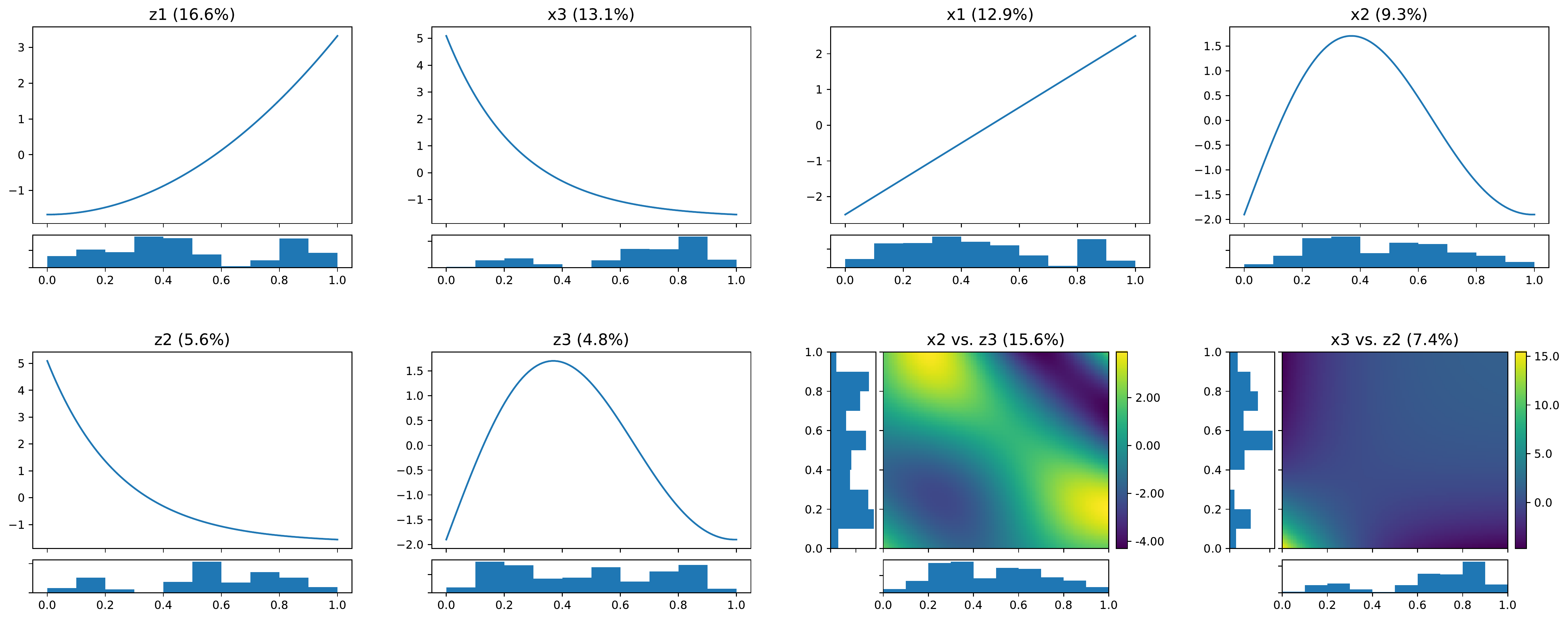}}
		\quad
		\caption{Main effects and manifest interactions of the simulation regression setting.}
		\label{simulation_gami}
	\end{figure}

	Fig.~\ref{simulation_gami} draws the global interpretation of main effects, manifest interactions of the simulation study (regression scenario). Note that in the original formulation, it is assumed to have 6 active main effects $\{x_1,  z_1\}$ and 2 active interaction effects $\{(x_2, z_3),(x_3, z_2)\}$. We can equivalently separate marginal effects from the interactions so that the active main effects also include $\{x_2,x_3,z_2,z_3\}$. It can be seen that GAMMLI captures all the important feature effects. 
	
\begin{figure}[!t]
		\centering
		\includegraphics[width=0.7\textwidth]{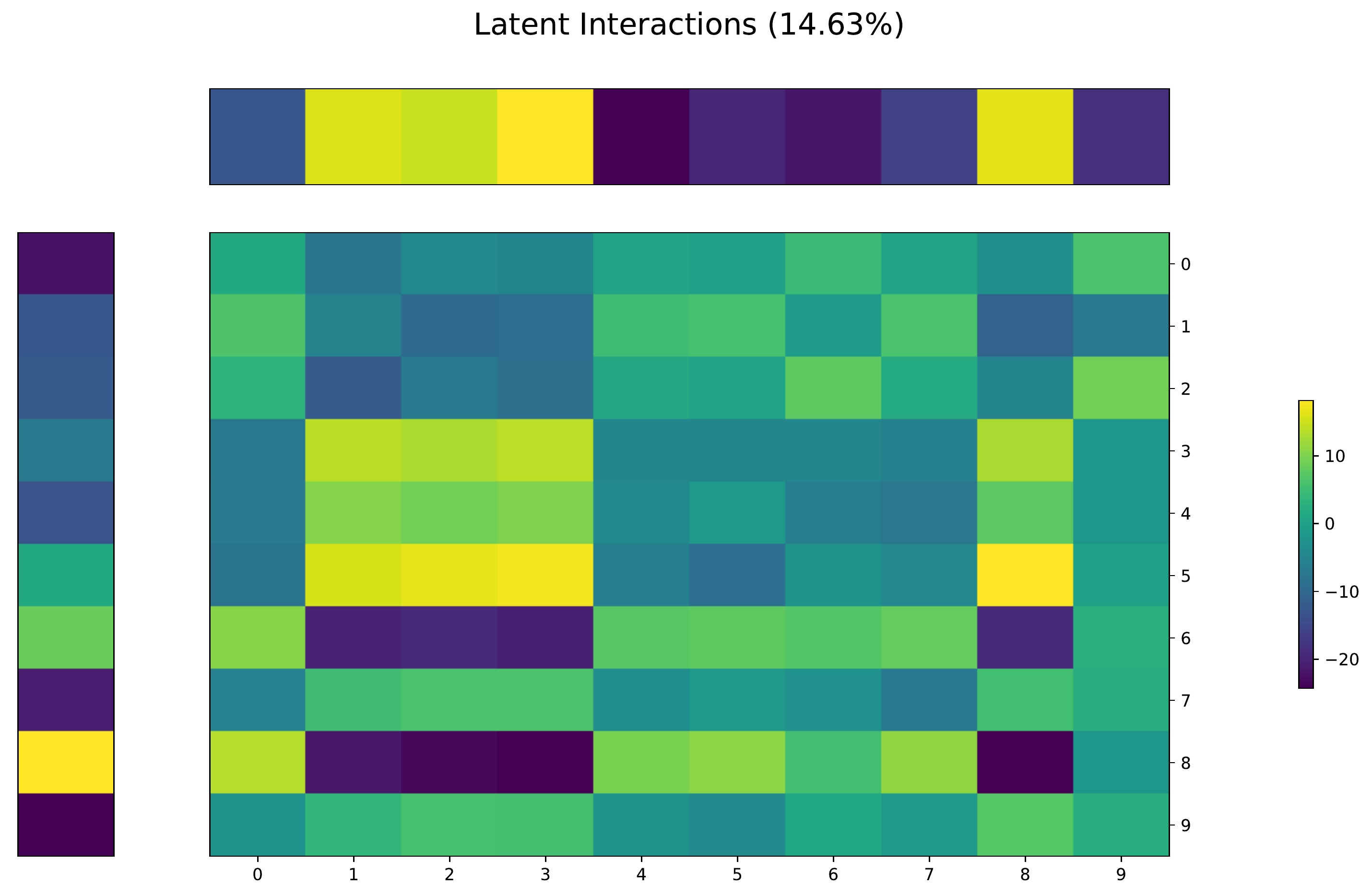}
		\caption{Global interpretation of latent interactions (simulation regression).}
		\label{simulation_latent}
	\end{figure}

	Fig.~\ref{simulation_latent} presents the global interpretation of latent interaction effects with the IR of 14.63\%. The user group number $K$ (the y-axis) and item group number $L$ (the x-axis) are set to 10. From Fig.~\ref{simulation_latent}, we can find that user groups 7 and 9 are most similar, as they have similar margin effects and similar preferences for different item groups. Moreover, user group 5 has the strongest preference of item group 8 among all group interactions.
	
\begin{figure}[htb!]
		\centering
		\centerline{\includegraphics[width=0.7\textwidth]{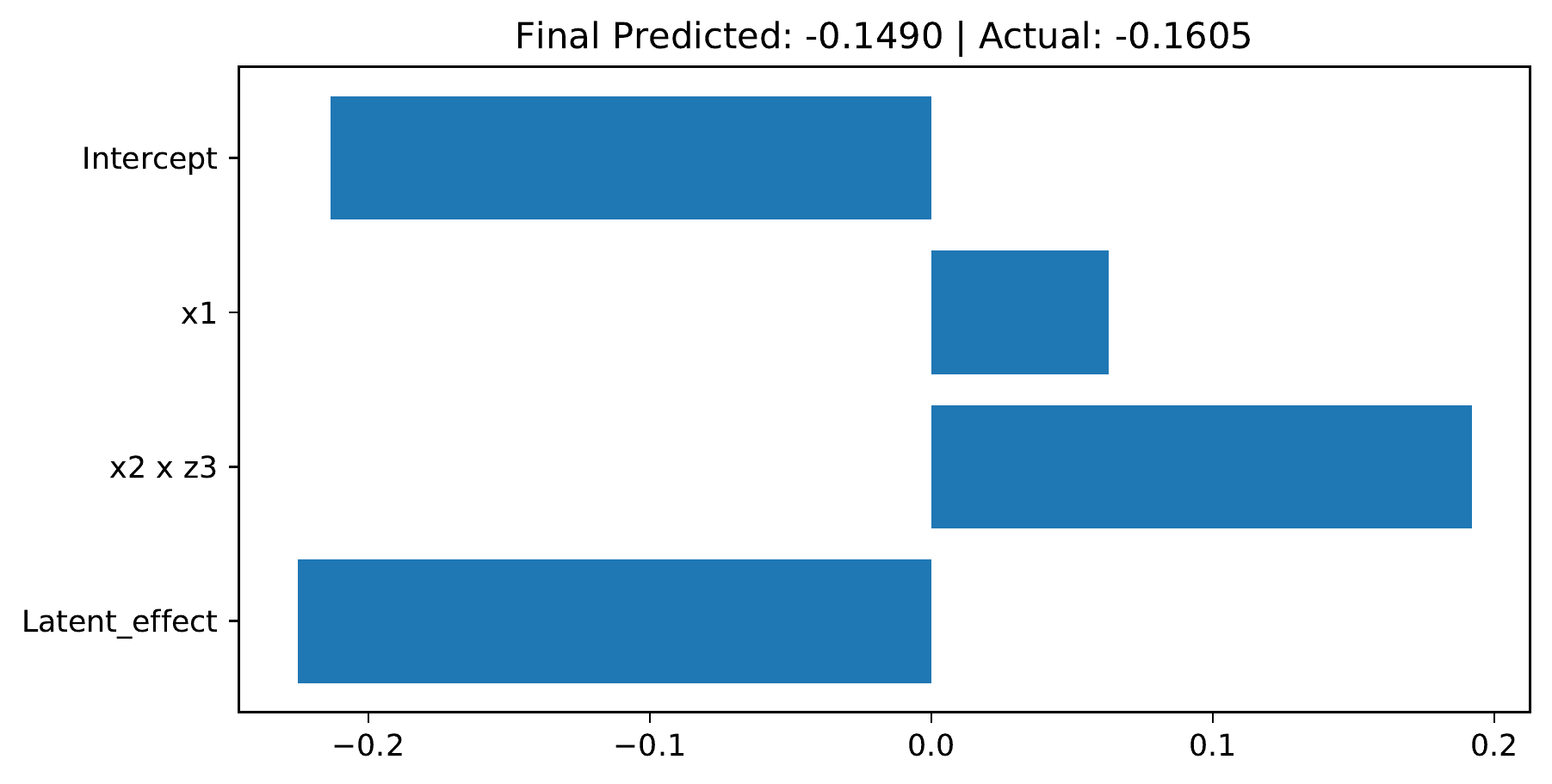} \ \ \  \ \ \ \ \ \ \ \ \ }
		\caption{Local interpretation for one randomly picked sample (simulation regression).}
		\label{simulatiuon_local}
	\end{figure}

	Except for the global interpretation, GAMMLI can also be locally explained, leading to a transparent personal recommendation system. Given a user and an item, the model outputs the final decision and why we recommend this item to this user. The additive components of main effects, manifest interactions, and latent interactions are provided. We can also get the local interpretation of the simulation study (regression scenario) from the prediction diagnosis of one sample point in Fig.~\ref{simulatiuon_local}. This information will be useful to understand the recommendation results.
	
	Table~\ref{tab:performance_comparison_simu} reports the average test accuracy of our model and several competitive black-box recommendation system models under both regression and classification scenarios. For each scenario, the best model is highlighted in bold. It can be observed that our proposed GAMMLI performs the best under both regression and classification scenario. While maintaining superior prediction accuracy, we achieve much better interpretability, which leads our model more transparent and understandable.
	
	\begin{table}[!t]
		\centering
		\begin{threeparttable}
			\caption{Testing accuracy comparison under simulation study.}
			\label{tab:performance_comparison_simu}
			\begin{tabular}{ccccc}
				\toprule
				\multirow{2}{*}{Method}&
				\multicolumn{2}{c}{ Regression}&\multicolumn{2}{c}{Classification}\\
				\cmidrule(lr){2-3} \cmidrule(lr){4-5}
				&MAE&RMSE&AUC&Log Loss\\
				\midrule
				GAMMLI&\textbf{0.3613}$\pm$0.0998&\textbf{0.5405}$\pm$0.1357&\textbf{0.9760}$\pm$0.0022& \textbf{0.1972}$\pm$0.0091\\
				FM&0.5239$\pm$0.0813&0.7208$\pm$0.1094& 0.9389$\pm$0.0115& 0.3599$\pm$0.0393\\
				DeepFM&0.4449$\pm$0.0194&0.6149$\pm$0.0292& 0.9437$\pm$0.0177& 0.3396$\pm$0.0566\\
				XGBoost&0.4574$\pm$0.0040&0.6360$\pm$0.0058& 0.9223$\pm$0.0003& 0.4121$\pm$0.0024\\
				SVD&1.6215$\pm$0.0000 &2.0697$\pm$0.0000& 0.7795$\pm$0.0000& 0.6610$\pm$0.0000\\
				\bottomrule
			\end{tabular}
		\end{threeparttable}
	\end{table}
	
	\subsection{Real Data Applications}
	We further consider three real-world datasets, MovieLens-1m dataset, the Santander bank dataset, and the University Recommendation dataset.
	
	\textbf{Movielens}~\footnote{\href{https://grouplens.org/datasets/movielens/1m/}{https://grouplens.org/datasets/movielens/1m/}} is a movie rating dataset that has been widely used to evaluate recommendation system algorithms. We used the version Movielens-1m, i.e., it contains one million ratings. Several features are taken into consideration, including gender, age, occupation, and genres. There are 6040 users and 3706 movies in this dataset, and the sparsity is 95.5\%.
	
	\textbf{Santander bank}~\footnote{\href{https://www.kaggle.com/c/santander-product-recommendation}{https://www.kaggle.com/c/santander-product-recommendation}} is an open Kaggle dataset. The dataset provides 1.5 years of customer behavior data from Santander bank to predict what new products the customers will purchase on 2016-06-28. In the Santander dataset, 22 bank products and 24 profile features of users are taken into consideration. The 22 bank products are further summarized into three categories, which form a new item feature, \textit{item property}. To remove the effect of time and make it convenient for modeling, we only collect samples where users have purchase behavior on 2016-06-28 as positive samples. We set the records that users without purchased items as the initial negative samples. To distinguish the real negative samples and the unknown samples, we first regard the negative samples as missing values and use the SoftImpute algorithm to fill the missing values. After imputing the missing values, we set a proper threshold to define reliable negative samples \citep{zhang2005simple}.
	
	\textbf{University Recommendation}~\footnote{\href{https://www.kaggle.com/nitishabharathi/university-recommendation}{https://www.kaggle.com/nitishabharathi/university-recommendation}} is an open Kaggle dataset. It contains over 14000 students' profiles in computer science-related fields with admits/rejects to 54 different universities in the USA. Universities' rankings and public status are manually collected from US News~\footnote{\href{https://www.usnews.com/best-colleges}{https://www.usnews.com/best-colleges}} as item features.
	
	\begin{table}[!t]
		\footnotesize
		\renewcommand\tabcolsep{2pt}
		\centering
		\begin{threeparttable}
			\caption{Testing accuracy comparison under real data.}
			\label{tab:performance_comparison_real}
			\begin{tabular}{ccccccc}
				\toprule
				\multirow{2}{*}{Method}&
				\multicolumn{2}{c}{ Movielens}&\multicolumn{2}{c}{Santander}&\multicolumn{2}{c}{University Recommendation}\\
				\cmidrule(lr){2-3} \cmidrule(lr){4-5}\cmidrule(lr){6-7}
				&MAE&RMSE&AUC&Log Loss&AUC&Log Loss\\
				\midrule
				GAMMLI& 0.7095$\pm$0.0047&0.9187$\pm$0.0038 & \textbf{0.9954}$\pm$0.0006 &\textbf{0.0463}$\pm$0.0026&0.7486$\pm$0.0064  &0.6304$\pm$0.0093\\
				FM&0.6755$\pm$0.0024   &0.8584$\pm$0.0013&0.9901$\pm$0.0016 &0.1025$\pm$0.0052&0.7046$\pm$0.0420 &0.7105$\pm$0.1155\\
				DeepFM&\textbf{0.6748}$\pm$0.0022 &\textbf{0.8578}$\pm$0.0008 & 0.9899$\pm$0.0020&0.1030$\pm$0.0066&0.7207$\pm$0.0288 &0.6801$\pm$0.0619\\
				XGBoost&0.8634$\pm$0.0000 &1.0622$\pm$0.0000& 0.9947$\pm$0.0002&0.0507$\pm$0.0010&\textbf{0.7780}$\pm$0.0047&\textbf{0.5647}$\pm$0.0048\\
				SVD&0.9041$\pm$0.0000  &1.1373$\pm$0.0000& 0.9867$\pm$0.0000 &0.0851$\pm$0.0000&0.5110$\pm$0.0000 &0.6924$\pm$0.0000\\
				\bottomrule
			\end{tabular}
		\end{threeparttable}
	\end{table}
	
	As shown in Table~\ref{tab:performance_comparison_real}, GAMMLI shows comparative predictive performance to that of other benchmark models. It can be observed that GAMMLI outperforms other benchmark models in the Santander dataset and keeps a small disparity with the most accurate benchmark models in Movielens and University Recommendation. Moreover, other benchmark models are not that stable among the three datasets. GAMMLI preserves relatively advantageous performance, which also shows its robustness. In practice, GAMMLI is more likely to have better predictive performance when the observed features have strong signals. While maintaining relatively superior prediction accuracy, a clear interpretation of GAMMLI is generated. We take the University Recommendation dataset as an example to demonstrate the interpretability of GAMMLI.
	
\begin{figure}[htb!]
		\centering
		\includegraphics[width=1.0\textwidth]{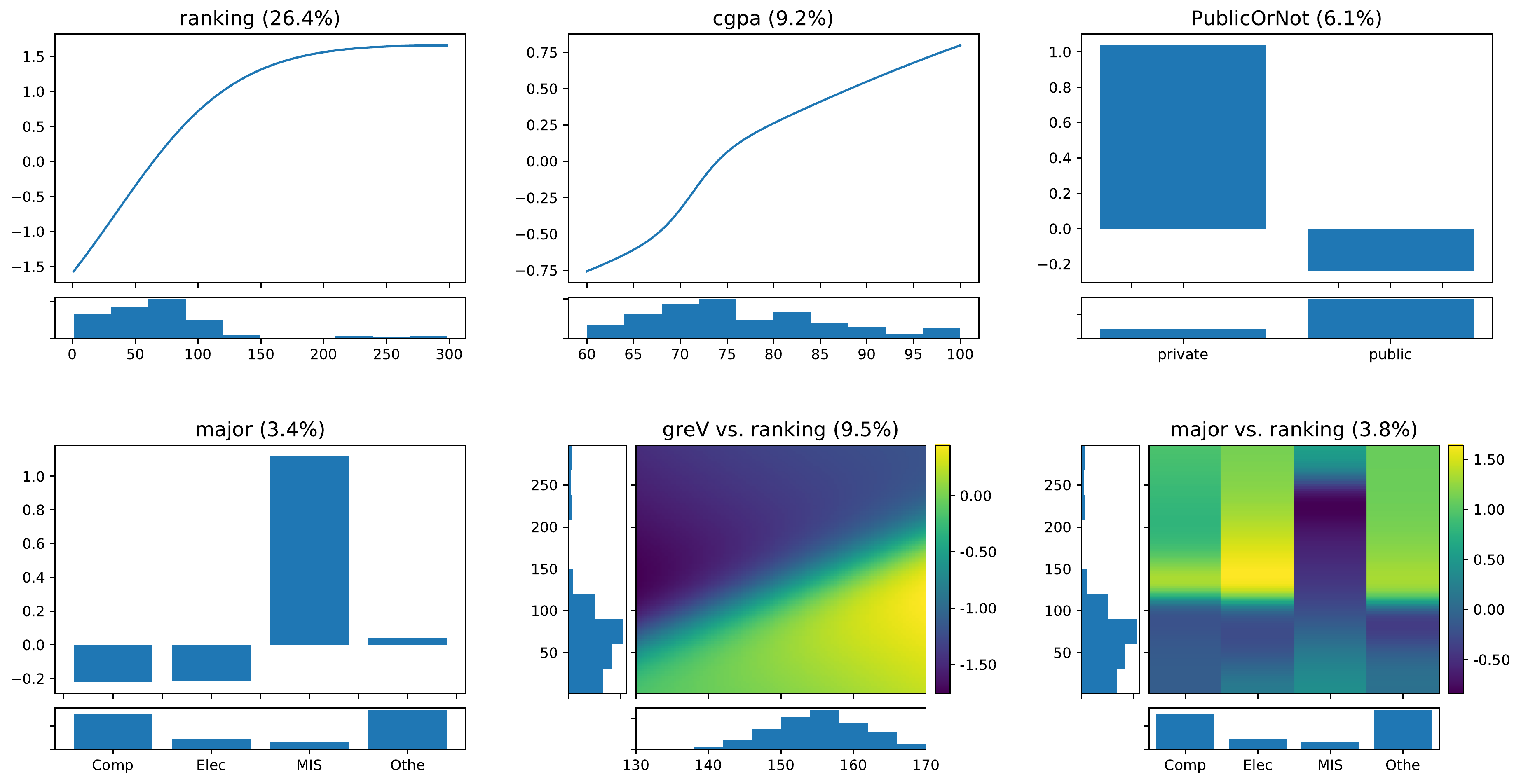}
		\caption{Global interpretation of main effects and manifest interactions (University Recommendation).}
		\label{university_maineffects}
	\end{figure}

	The global interpretation of the main effects of the University Recommendation dataset is shown in Fig.~\ref{university_maineffects}. From the univariate plots, we can get four significant main effects in University Recommendation which are \textit{ranking} (University US ranking), \textit{cgpa}, \textit{PublicOrNot}, \textit{major} with IRs equal to 26.4\%, 9.2\%, 6.1\%, 3.4\%, respectively. In addition to the univariate plots, the density plots of the corresponding variables are also provided. The most significant item feature is \textit{ranking}. It can be observed that students would have a higher chance to be admitted if they apply to universities with lower rankings. For those top universities, though many students submit their applications, most of them got rejected. Another significant item feature is \textit{PublicOrNot}. The admission rate of private universities is around 60\%, while public universities' admission rate is only around 50\%. The most significant user feature is \textit{cgpa}. The high score of cumulative GPA has a monotonically positive impact on admission decisions. 
	
	Regarding manifest interactions, it can be observed that two manifest interactions are captured, which are \textit{greV vs. ranking} and \textit{major vs. ranking} with IRs equal to 9.5\% and 3.8\%. As the main effects are removed due to the marginal clarity constraint, the estimated pairwise interactions can be individually illustrated. The most significant manifest interactions is \textit{greV vs. ranking}. For the students with GRE verbal higher than around 155, the students may be more likely to be admitted by the Top 50 universities. In practice, for students major in computer science-related subjects, most of them are skilled in mathematics and get high scores in the GRE quantitative test. A high GRE verbal score might be equivalent to a high GRE score.
	
\begin{figure}[htb!]
		\centering
		\includegraphics[width=0.7\textwidth]{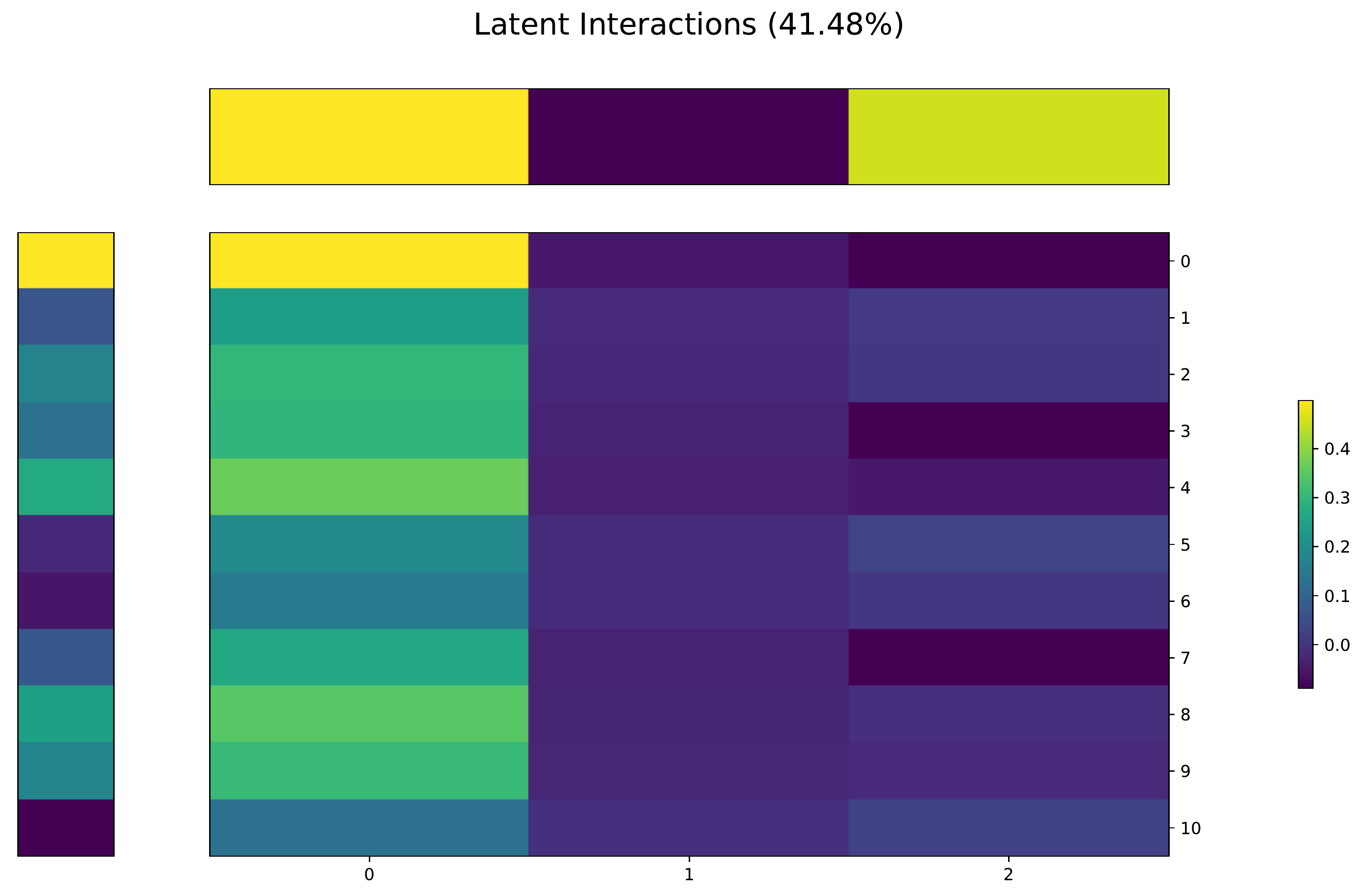}
		\caption{Global interpretation of latent interactions (University Recommendation).}
		\label{university_latent}
	\end{figure}

\begin{figure}[htb!]
		\centering
		\includegraphics[width=0.5\textwidth]{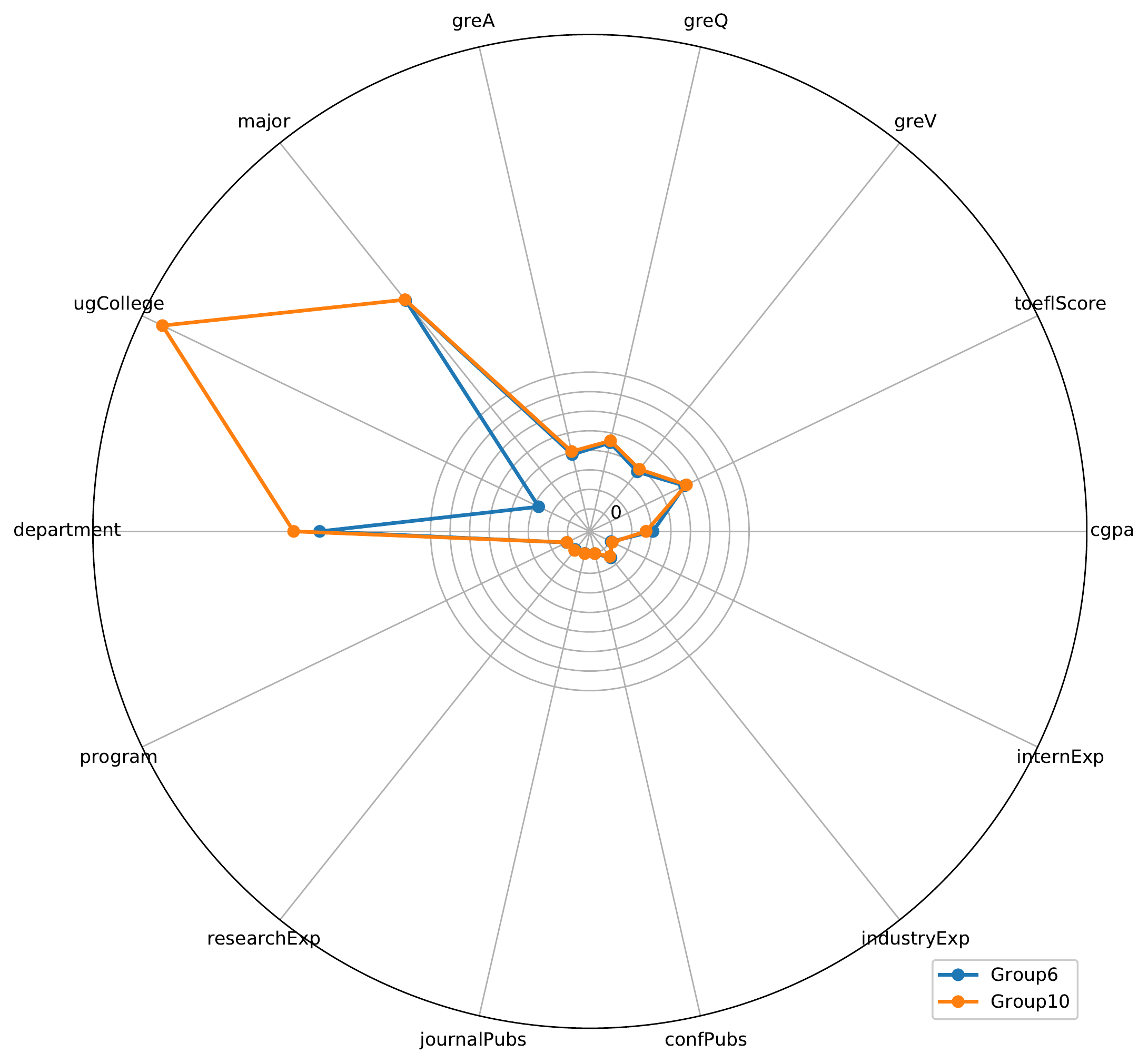}
		\caption{Radar plot for user groups 6 and 10 (University Recommendation).}
		\label{university_radar}
	\end{figure}

	Fig.~\ref{university_latent} shows the global interpretation of latent interaction effects are significantly important with the IR equals to 41.48\%. The user group number $K=11$ and item group number $L=3$ are chosen by the validation set. It can be observed that user groups 6 and 10 are most similar among all user groups for they have similar sidebar colors and similar preferences for different item groups. The strongest preference is obtained between user group 0 and item group 0 among all latent interactions. We use radar plot Fig.~\ref{university_radar} to portray the centroids of similar groups in terms of their profiles. From fig.~\ref{university_radar}, we can find that the difference between user group 6 and user group 10 mainly attributes to \textit{ugCollege} which refers to as applicants' undergraduate college. Note \textit{ugCollege} is not an important feature from the analysis of main and manifest interactions. 
	
	\subsection{Cold Start Problem}
	When a user or item is input into the system, it is automatically judged whether it is new. As for a regular user or item, the system will assign the existing latent vectors to it. If it is a new one, only the user/item features are given, and there will be no existing latent vectors. Therefore, we can find its nearest feature group and assign it with the group's centroid of latent factors. Finally, the results of all three stages are combined to provide recommendation results without any historical information. We test the performance of the cold start scenario by evaluating the test samples related to the new users or items under both simulation and University Recommendation datasets. The new users or items denote the ones that do not appear in the training dataset but appear in the test dataset. Table~\ref{tab:performance_comparison_cold} shows that the performance of GAMMLI exceeds benchmark models by a great margin.
	
	\begin{table}[!t]
		\renewcommand\tabcolsep{2pt}
		\centering
		\begin{threeparttable}
			\caption{Cold start testing accuracy comparison.}
			\label{tab:performance_comparison_cold}
			\begin{tabular}{ccccc}
				\toprule
				\multirow{2}{*}{Method}&
				\multicolumn{2}{c}{Simulation}&\multicolumn{2}{c}{University Recommendation}\\
				\cmidrule(lr){2-3} \cmidrule(lr){4-5}
				&MAE&RMSE&AUC&Log Loss\\
				\midrule
				GAMMLI&\textbf{1.3196}$\pm$0.4966 &\textbf{1.5130}$\pm$0.4029 & \textbf{0.7301}$\pm$0.0185&\textbf{0.5727}$\pm$0.0647 \\
				FM&1.9265$\pm$0.6417 &2.3353$\pm$0.7770&0.6536$\pm$0.0382&0.6856$\pm$0.2656\\
				DeepFM&2.1997$\pm$0.5650 &2.5896$\pm$0.4705&0.6654$\pm$0.0190&0.6113$\pm$0.0415\\
				XGBoost&1.3400$\pm$0.2624 &1.5200$\pm$0.2568&0.7169$\pm$0.0173&0.5964$\pm$0.0133\\
				SVD&3.6663$\pm$0.0000 &4.1352$\pm$0.0000&0.6173$\pm$0.0000 &0.7059$\pm$0.0000\\
				\bottomrule
			\end{tabular}
		\end{threeparttable}
	\end{table} 
	
	\section{Conclusion} \label{conclusion}
	In this paper, a novel intrinsically explainable recommendation system called GAMMLI is proposed based on the generalized additive model with manifest interactions and latent interactions. It captures the main effects and manifest interaction of observed features and discovers the latent interactions of unobserved features. All effects can be interpreted by visualization methods, including 1-D line plots and 2-D heatmaps. The experiment results show that GAMMLI has advantages in both predictive performance and interpretability. Meanwhile, the experiment results support that GAMMLI can tackle cold start problems.	
	
	\bibliographystyle{apalike}
	\bibliography{Reference}
	
\end{document}